\documentclass[conference]{IEEEtran}
\usepackage{times}

\usepackage[numbers]{natbib}
\usepackage{multicol}
\usepackage[bookmarks=true]{hyperref}
\usepackage{xcolor}
\usepackage{listings}

\newcommand\myshade{85}
\colorlet{mylinkcolor}{violet}
\colorlet{mycitecolor}{orange}
\colorlet{myurlcolor}{blue}

\definecolor{dkgreen}{rgb}{0,0.6,0}
\definecolor{gray}{rgb}{0.5,0.5,0.5}
\definecolor{mauve}{rgb}{0.58,0,0.82}

\hypersetup{
  linkcolor  = mylinkcolor!\myshade!black,
  citecolor  = mycitecolor!\myshade!black,
  urlcolor   = myurlcolor!\myshade!black,
  colorlinks = true,
  linktoc = none,
}

\IEEEoverridecommandlockouts                              % This command is only needed if 
\usepackage{amsmath,amsfonts,bm}

\def\eqref#1{equation~\ref{#1}}
\def\1{\bm{1}}

\DeclareMathAlphabet{\mathsfit}{\encodingdefault}{\sfdefault}{m}{sl}
\SetMathAlphabet{\mathsfit}{bold}{\encodingdefault}{\sfdefault}{bx}{n}

\usepackage{graphicx}
\usepackage{amsmath}
\usepackage{algpseudocode}
\usepackage{algorithm}
\usepackage[algo2e]{algorithm2e}
\usepackage{multirow}
\usepackage[normalem]{ulem}
\usepackage{booktabs}
\usepackage{graphicx}
\newcommand{\xxnote}[3]{}
\ifx\hidenotes\undefined
  \renewcommand{\xxnote}[3]{\color{#2}{#1: #3}}
\fi

\hypersetup{
    colorlinks=true,
    linkcolor=blue,
    filecolor=magenta,      
    urlcolor=blue,
}

\title{\LARGE \bf The Surprising Effectiveness of Representation Learning \\for Visual Imitation}
\author{%
\\
Jyothish Pari$^\ast$\\
New York University\\
\texttt{jp5981@nyu.edu}
\and
Nur Muhammad\\(Mahi) Shafiullah$^\ast$\\
New York University\\
\texttt{mahi@cs.nyu.edu}
\and
Sridhar Pandian\\Arunachalam\\
New York University\\
\texttt{sa5914@nyu.edu}
\and
\\
Lerrel Pinto \\
New York University\\
\texttt{lerrel@cs.nyu.edu}
}%

\begin{document}
\maketitle
\thispagestyle{empty}
\pagestyle{empty}

\def\thefootnote{*}\footnotetext{The first two authors contributed equally to this work.}\def\thefootnote{\arabic{footnote}}
%%%%%%%%%%%%%%%%%%%%%%%%%%%%%%%%%%%%%%%%%%%%%%%%%%%%%%%%%%%%%%%%%%%%%%%%%%%%%%%%
\begin{abstract}
While visual imitation learning offers one of the most effective ways of learning from visual demonstrations, generalizing from them requires either hundreds of diverse demonstrations, task specific priors, or large, hard-to-train parametric models.
One reason such complexities arise is because standard visual imitation frameworks try to solve two coupled problems at once: learning a succinct but good representation from the diverse visual data, while simultaneously learning to associate the demonstrated actions with such representations.
Such joint learning causes an interdependence between these two problems, which often results in needing large amounts of demonstrations for learning.
To address this challenge, we instead propose to decouple representation learning from behavior learning for visual imitation. First, we learn a visual representation encoder from offline data using standard supervised and self-supervised learning methods. Once the representations are trained, we use non-parametric Locally Weighted Regression to predict the actions.
We experimentally show that this simple decoupling improves the performance of visual imitation models on both offline demonstration datasets and real-robot door opening compared to prior work in visual imitation. All of our generated data, code, and robot videos are publicly available at \url{https://jyopari.github.io/VINN/}.
\end{abstract}

%%%%%%%%%%%%%%%%%%%%%%%%%%%%%%%%%%%%%%%%%%%%%%%%%%%%%%%%%%%%%%%%%%%%%%%%%%%%%%%%
\section{Introduction}
% \lpnote{TODO: Lerrel}

Imitation learning serves as a powerful framework for getting robots to learn complex skills in visually rich environments~\cite{zhang2018deep, stadie2017third,duan2017one,zhu2018reinforcement,young2020visual}. 
Recent works in this area have shown promising results in generalization to previously unseen environments for robotic tasks such as pick and place, pushing, and rearrangement~\cite{young2020visual}. 
However, such generalization is often too narrow to be directly applied in the diverse real-world application. 
For instance, policies trained to open one door rarely generalize to opening different doors~\cite{DBLP:journals/corr/abs-1908-01887}.
This lack of generalization is further exacerbated by the plethora of different options to achieve generalization: either needing hundreds of diverse demonstrations, task-specific priors, or large parametric models. 
This begs the question: What really matters for generalization in visual imitation?

An obvious answer is visual representation -- generalizing to diverse visual environments should require powerful representation learning. 
Prior work in computer vision~\cite{byol,simclr,moco2,swav,bardes2021vicreg} have shown that better representations significantly improve downstream performance for tasks such as image classification.
However, in the case of robotics, evaluating the performance of visual representations is quite complicated.
Consider behavior cloning~\cite{torabi2018behavioral}, one of the simplest methods of imitation. 
Standard approaches in behavior cloning fit convolutional neural networks on a large dataset of expert demonstrations using end-to-end gradient descent.
Although powerful, such models conflate two fundamental problems in visual imitation: (a) representation learning, i.e. inferring information-preserving low-dimensional embeddings from high-dimensional observations and (b) behavior learning, i.e. generating actions given representations of the environment state. This joint learning often results in large dataset requirements for such techniques. 

\begin{figure}[t]
  \begin{center}
    \includegraphics[width = \linewidth]{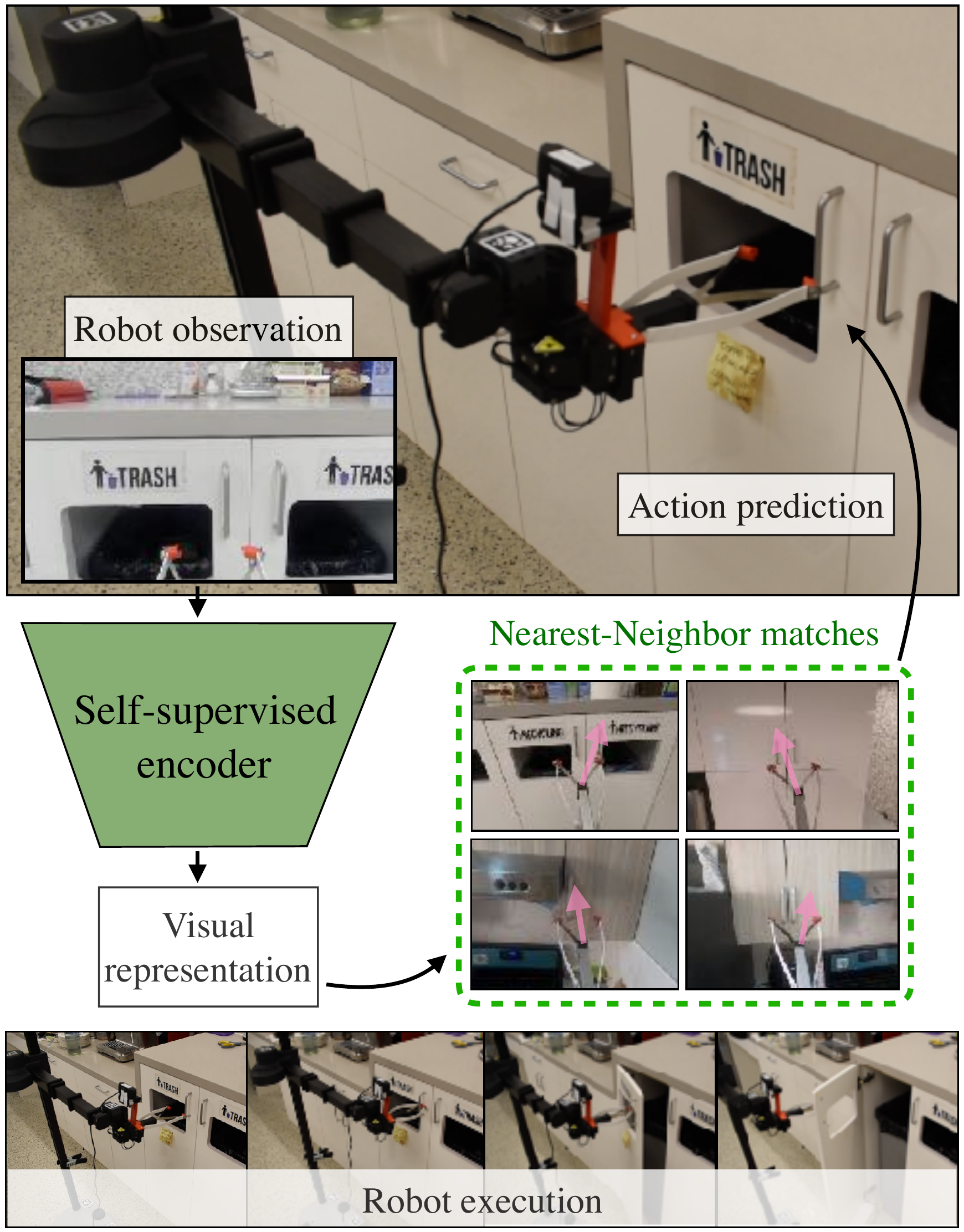}
  \end{center}
  \vspace{-0.1in}
  \caption{Consider the task of opening doors from visual observations. VINN, our visual imitation framework first learns visual representations through self-supervised learning. Given these representations, non-parametric weighted nearest neighbors from a handful of demonstrations is used to compute actions, which results in robust door-opening behavior.}
  \vspace{-0.1in}
\label{fig:intro}
\end{figure}

One way to achieve this decoupling is to use representation modules pre-trained through standard proxy tasks such as image classification, detection, or segmentation~\cite{sax2019learning}.
However, this relies on large amounts of labelled human data on datasets that are often significantly out of distribution to robot data~\cite{chen2020robust}.
A more scalable approach is to take inspiration from recent work in computer vision, where visual encoders are trained using self-supervised losses~\cite{moco2, simclr, byol}.
These methods allow the encoders to learn useful features of the world without requiring human labelling.
There has been recent progress in vision-based Reinforcement Learning (RL) that improves performance by creating this explicit decoupling \cite{stooke2021decoupling, yarats2021reinforcement}.
Visual imitation has a significant advantage over RL settings: learning visual representations in RL is further coupled with challenges in exploration~\cite{yarats2021mastering}, which has limited its application in real-world settings due to poor sample complexity. 

In this work we present a new and simple framework for visual imitation that decouples representation learning from behavior learning. First, given an offline dataset of experience, we train visual encoders that can embed high-dimensional visual observations to low-dimensional representations. Next, given a handful of demonstrations, for a new observation, we find its associated nearest neighbors in the representation space. For our agent's behavior on that new observation, we use a weighted average of the nearest neighbors' actions. This technique is inspired by Locally Weighted Regression~\cite{atkeson1997locally}, where instead of operating on state estimates, we operate on self-supervised visual representations. Intuitively, this allows the behavior to roughly correspond to a Mixture-of-Experts model trained on the visual demonstrations. Since nearest neighbors is non-parametric, this technique requires no additional training for behavior learning. We will refer to our framework as Visual Imitation through Nearest Neighbors (VINN).

Our experimental analysis demonstrates that VINN can successfully learn powerful representations and behaviors across three manipulation tasks: Pushing, Stacking, and Door Opening. 
Surprisingly, we find that non-parametric behavior learning on top of learned representations is competitive with end-to-end behavior cloning methods. On offline MSE metrics, we report results on par with competitive baselines, while being significantly simpler. 
To further test the real-world applicability of VINN, we run robot experiments on opening doors using 71 visual demonstrations. Across a suite of generalization experiments, VINN succeeds 80\% on doors present in the demonstration dataset and 40\% on opening the door in novel scenes. 
In contrast, our strongest baselines have success rates of 53.3\% and 3.3\% respectively.

To summarize, this paper presents the following contributions. 
First, we present VINN, a novel yet simple to implement visual imitation framework that derives non-parametric behaviors from learned visual representations. 
Second, we show that VINN is competitive to standard parametric behavior cloning and can outperform it on a suite of manipulation tasks. Third, we demonstrate that VINN can be used on real robots for opening doors and can achieve high generalization performance on novel doors. Finally, we extensively ablate over and analyze different representations, amount of training data, and other hyperparameters to demonstrate the robustness of VINN.
\section{Related Work}

\begin{figure*}[h]
  \begin{center}
    \includegraphics[width = 1.0\textwidth]{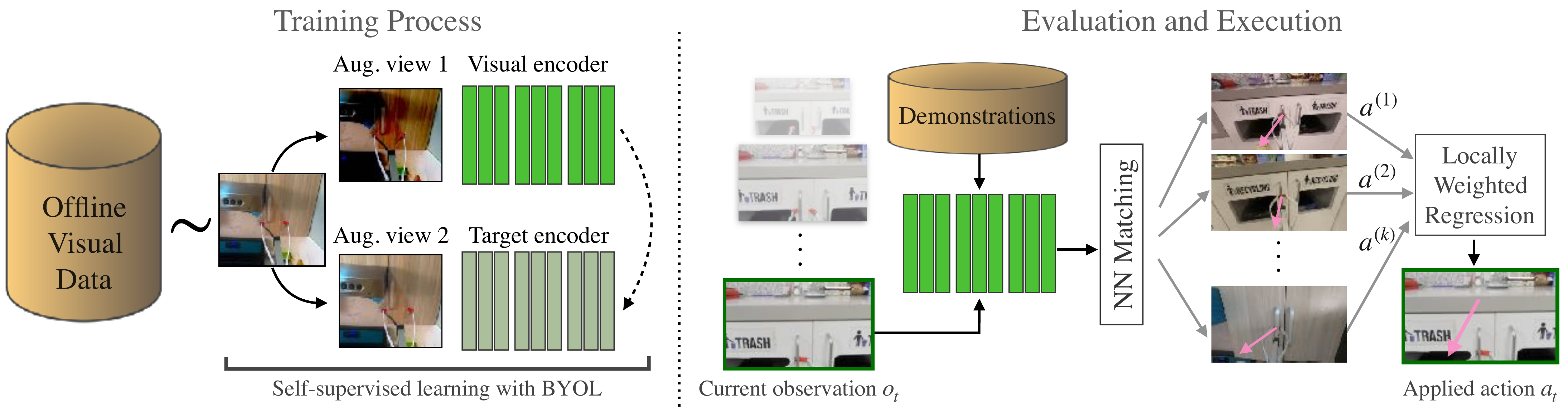}
  \end{center}
  \caption{Overview of our VINN algorithm. During training, we use offline visual data to train a BYOL-style self-supervised model as our encoder. During evaluation, we compare the encoded input against the encodings of our demonstration frames to find the nearest examples to our query. Then, our model's predicted action is just a weighted average of the associated actions from the nearest images.}
\label{fig:arch}
\end{figure*}

\subsection{Imitation via Cloning}
Imitation learning is frequently used to learn skills and behaviors from human demonstrations~\cite{piaget2013play, meltzoff1977imitation, meltzoff1983newborn, tomasello1993imitative}. 
In the context of manipulation, such techniques have successfully solved a variety of problems in pushing, stacking, and grasping~\cite{zhang2018deep,DBLP:journals/corr/abs-1802-09564,argall2009survey,hussein2017imitation}. Behavioral Cloning (BC) \cite{torabi2018behavioral} is one of the most common techniques. 
% A comprehensive review of imitation learning can be found in \cite{argall2009survey, hussein2017imitation}.
If the agent's morphology or viewpoint is different than the demonstrations', the model needs to involve techniques such as transfer learning to resolve this domain gap~\cite{stadie2017third,sermanet2016unsupervised}. 
To close this unintended domain gap, \cite{zhang2018deep} has used tele-operation methods, while \cite{song2020grasping, young2020visual} have used assistive tools. 
Using assistive tools provides us the benefit of being a able to scalably collect diverse demonstrations. In this paper, we follow the DemoAT \cite{young2020visual} framework to collect expert demonstrations.

\subsection{Visual Representation Learning}

In computer vision, interest in learning a good representation has been longstanding, especially when labelled data is rare or difficult to collect \cite{simclr, moco2, byol, swav}.
This large class of representation learning techniques aim to extract features that can help other models improve their performance in some downstream learning tasks, without needing to explicitly learn a label.
In such tasks, first a model is trained on one or more pretext tasks with this unlabeled dataset to learn a representation.
Such tasks generally include instance invariance, or predicting some image transformation parameters (e.g. rotation and distortion), patches, or frame sequence~\cite{gidaris2018unsupervised,dosovitskiy2015discriminative,doersch2016unsupervised,misra2016shuffle,simclr,moco2,wu2018unsupervised}.
% Some works have proposed simultaneously training these pretext tasks alongside the main objective~\cite{zhai2019s4l, sun2019unsupervised}.
In representation learning, the performance of the model on the pretext task is usually disregarded.
Instead, the focus is on
% the model pretrained through these pretext tasks and 
the input domain to representation mapping that these models have learned.
Ideally, to solve such pretext tasks, the pretrained model may have learned some useful structural meaning and encoded it in the representation.
Thus, intuitively, such a model can be used in downstream tasks where there is not enough data to learn this structural meaning directly from the available task-relevant data.
Unsupervised representation learning, in works such as \cite{simclr, moco2, byol, swav, bardes2021vicreg, dwibedi2021little}, has shown impressive performance gains on difficult benchmarks since they can harness a large amounts of unlabelled data unavailable in task-specific datasets.

Recently, interest in unsupervised or semi-supervised representation learning technique has grown within robotics~\cite{manuelli2020keypoints} due to the availability of unlabeled data and its effectiveness in visual imitation tasks ~\cite{young2021playful,zhan2020framework}.
We follow a BYOL-style~\cite{byol} self-supervised representation learning framework in our experiments.

\subsection{Non-parametric Control}
Non-parametric models are those, which instead of modeling some parameters about the data distribution, tries to express it in terms of previously observed training data. Non-parametric models are significantly more expressive, but as a downside to this, they usually require a large number of training examples to generalize well. A popular and simple example of non-parametric models is Locally Weighted Learning (LWL)~\cite{atkeson1997locally}. LWL is a form of instance-based, non-parametric learning that refers to algorithms whose response to any query is a weighted aggregate of similar examples. 
% There, similarity is measured in distance in some representation space.
Simple nearest neighbor models are an example of such learning, where all weight is put on the closest neighbor to the input point. Nearest neighbor methods have been successfully used in previous works for control tasks ~\cite{mansimov2018simple}
More sophisticated, $k$-NN algorithms base their predictions on an aggregate of the nearest $k$ points ~\cite{10.1007/978-1-4612-2660-4_33}.

Uses of LWL based methods in supervised learning, robotics, and reinforcement learning is quite old.
In works like~\cite{snell2017prototypical, wang2019simpleshot}, effectiveness of LWL algorithms like k-nearest neighbor has shown competitive success in difficult, high dimensional tasks like classifying the miniImageNet.
LWL has also shown success for robotic control problems~\cite{atkeson1997locally}, although it requires an accurate state-estimator to obtain low-dimensional states.
In~\cite{lee2016robust,pritzel2017neural,rajeswaran2018generalization}, elements of non-parametric learning is weaved into the reinforcement learning algorithms to create models which can adjust their complexity based on the amount of available data.
Finally, in works like~\cite{shah2018qlearning} non-parametric k-Nearest Neighbor regression based Q-functions are shown to give a good approximation of the true Q function under some theoretical assumptions.
Our work, VINN, draws inspiration from the simplicity of LWL and demonstrates the usefulness of this idea by using Locally Weighted Regression in challenging visual robotic tasks.

% In this work, we use a form of Locally Weighted regression with a weighting function~\cite{atkeson1997locally} to compute the particular action for an observation given similar observations among our demonstrations. 

%%%%%%%%%%%%%%%%%%%%%%%%%%%%%%%%%%%%%%%%%%%%%%%%%%%%%%%%%%%%%%%%%%%%%%%%%%%%%%%%
% \section{Background}
% I dont think we need background. We can describe high level in related work.

% \subsection{Representation Learning through Bootstrapping}

% \subsection{Weighted Nearest Neighbours}
\section{Approach}
% \lpnote{TODO: Mahi}

In this section, we describe the components of our algorithms and how they fit together to create VINN. As seen in Fig.~\ref{fig:arch}, VINN consists of two parts: (a) training an encoding network on offline visual data, and (b) querying against the provided demonstrations for a nearest-neighbor based action prediction. 

\subsection{Visual Representation Learning}

Given an offline dataset of visual experience from the robot, we first learn a visual representation embedding function.
In this work, we use two key insights for learning our visual representation: first, we can learn a good vision prior using existing large but unrelated real world datasets, and then, we can fine-tune starting from that prior using our demonstration dataset, which is small but relevant to the task at hand.

For the first insight, whenever possible, we initialize our models from an ImageNet-pretrained model. Such models are provided with the PyTorch~\cite{paszke2019pytorch} library that we use and can be achieved by simply adding a single parameter to the model initialization function call. 

Then, we use self supervised learning and train this visual encoder on the all the frames in our offline training dataset.
In this work, we use Bootstrap Your Own Latent (BYOL)~\cite{byol} as the self-supervision objective.
As illustrated in Fig.~\ref{fig:arch}, BYOL uses two versions of the same encoder network: one normally updating online network, and a slow moving average of the online network called the target network.
The BYOL self-supervised loss function tries to reduce the discrepancy in the two heads of the network when they are fed with differently augmented version of the same image.
Although we use BYOL in this work, VINN can also work with other self-supervised representation learning methods~\cite{simclr,moco2, swav,bardes2021vicreg} (Table~\ref{tab:mse-table}).

In practice, we initialize both the BYOL online and target networks with an ImageNet-pretrained encoder. 
Then, using the BYOL objective, we finetune them to better fit our image distribution.
Once the self-supervised training is done, we encode all our training demonstration frames with the encoder to obtain a set of their embeddings, $E$.

\subsection{$k$-Nearest Neighbors Based Locally Weighted Regression}
\begin{figure*}[ht]
  \begin{center}
    \includegraphics[width = \linewidth]{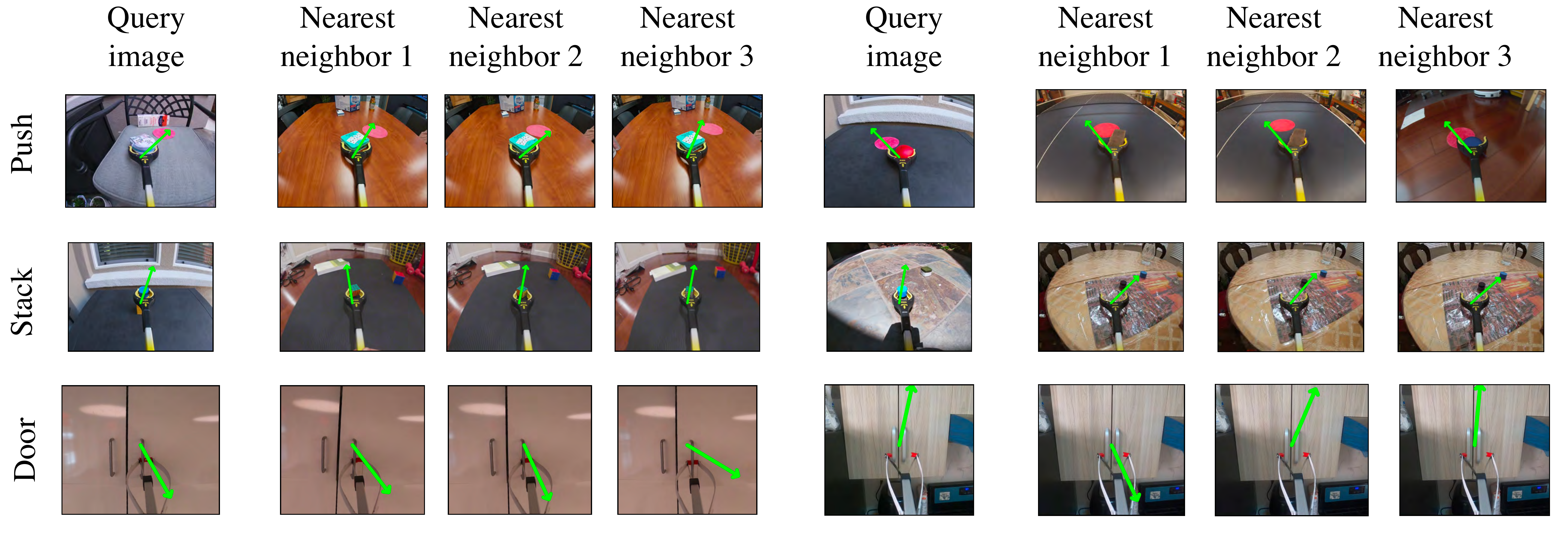}
  \end{center}
  \caption{Nearest neighbor queries on the encoded demonstration dataset; the query image is on the first column, and the found nearest neighbors are on the next three columns. The associated action is shown with a green arrow. The bottom right set of nearest neighbors demonstrates the advantage of performing a weighted average over nearest neighbors' actions instead of copying the nearest neighbor's action.}
\label{figure:sanity_check}
\end{figure*}

The set of embeddings $E$ given by our encoder holds compact representations of the demonstration images.
Thus, during test time, given an input we search for demonstration frames with similar features.
We find the nearest neighbors of the encoded input $e$ from the set of demonstration embeddings, $E$.
In Fig.~\ref{figure:sanity_check}, we see that these nearest neighbors are visually similar to the query image.
Our algorithm implicitly assumes that a similar observation must result in a similar action. 
Thus, once we have found the $k$ nearest neighbors of our query, we set the next action as an weighted average of the actions associated with those $k$ nearest neighbors.

Concretely, this is done by performing nearest neighbors search based on the distance between embeddings: $\|e - e^{(i)}\|_2$, where $e^{(i)}$ is the $i^{th}$ nearest neighbor. Once we find the $k$ nearest neighbors and their associated actions, namely $(e^{(1)}, a^{(1)}), (e^{(2)}, a^{(2)}), \cdots, (e^{(k)}, a^{(k)})$, we set the action as the Euclidean kernel weighted average \cite{atkeson1997locally} of those examples' associated actions:
% \begin{equation}
\[\hat a = \frac{\sum_{i=1}^k \exp \left ( {- \|e - e^{(i)}\|_2}\right ) \cdot a^{(i)} } {\sum_{i=1}^k \exp \left ( {- \|e - e^{(i)}\|_2}\right )}\]
% \end{equation}
In practice, this turns out to be the average of the observations' associated actions weighted by the SoftMin of their distance from the query image in the embedding space.

% \subsection{Representation Learning for Visual Imitation}

\subsection{Deployment in real-robot door opening}
\label{sec:demoat}
% \lpnote{Lerrel: Maybe this should go into experimental details?}
% \lpnote{Maybe this can involve the changes we made to the original VIME setup. Like the gripper type, camera, action extraction?}
For our robotic door opening task, we collect demonstrations using the DemoAT~\cite{young2020visual} tool.
Here, a reacher-grabber is mounted with a GoPro camera to collect a video of each trajectory.
We pass the series of frames into a structure from motion (SfM) method which outputs the camera's location in a fixed frame ~\cite{ozyesil2017survey}.
From the sequence of camera poses, which consist of coordinate and orientation, we extract translational motion which becomes our action.
To extract the gripper state, we train a gripper network that outputs a distribution over four classes (open, almost open, almost closed, closed), which represent various stages of gripping.
Then, we feed 
these images and their corresponding actions into our imitation learning method.

To train our visual encoders, we train ImageNet-pretrained BYOL encoders on individual frames in our demonstration dataset without action information. This same dataset with action information serves as the demonstration dataset for the $k$-NN based action prediction. Note that although we use task-specific demonstrations for representation learning, our framework is compatible with using other forms of unlabelled data such offline datasets~\cite{gulcehre2020rl,fu2020d4rl} or task-agnostic play data~\cite{young2021playful}.

To execute our door-opening skill on the robot, we run our model on a closed loop manner. After resetting the robot and the environment, on every step, we retrieve the robot observation and query the model with it. The model returns a translational action $\hat a$ as well as the gripper state $g$, and the robot moves $c \odot \hat a$ where the vector $c$ is a hyper-parameter with each element $< 1$ to mitigate our SfM model's inaccuracies and improve transfer from human demonstrations to robot execution. In addition, for nearest neighbor based methods, we have hyper-parameters that map the floating value $g$ into a gripper state which was tuned per experiment. 

%%%%%%%%%%%%%%%%%%%%%%%%%%%%%%%%%%%%%%%%%%%%%%%%%%%%%%%%%%%%%%%%%%%%%%%%%%%%%%%%
\section{Experimental Evaluation}\label{sec:experiments}
% \lpnote{TODO: Mahi}

\begin{figure*}[!htb]
\minipage{0.33\textwidth}
   \includegraphics[width=\linewidth]{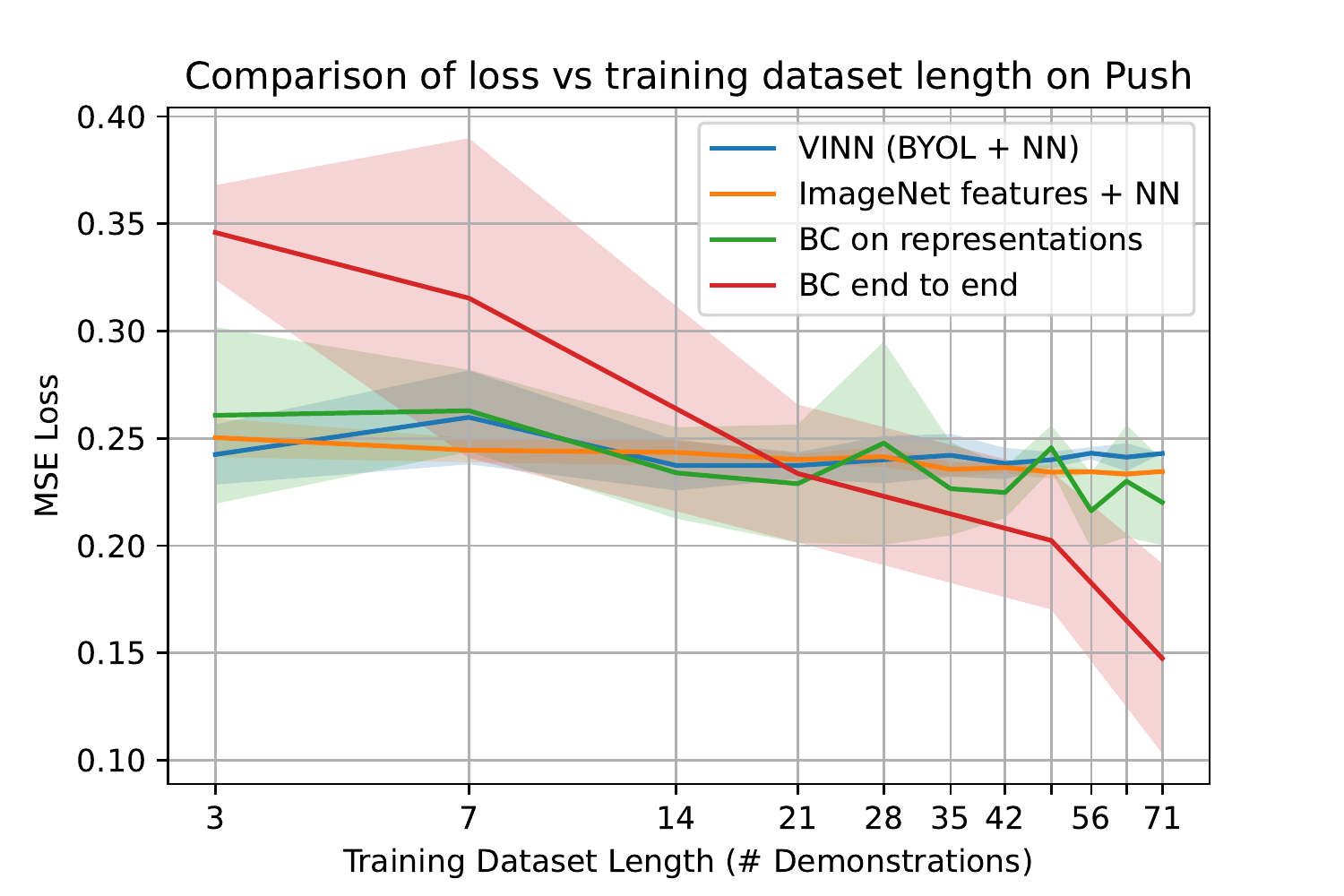}
  %\caption{A really Awesome Image}\label{fig:awesome_image1}
\endminipage\hfill
\minipage{0.33\textwidth}
  \includegraphics[width=\linewidth]{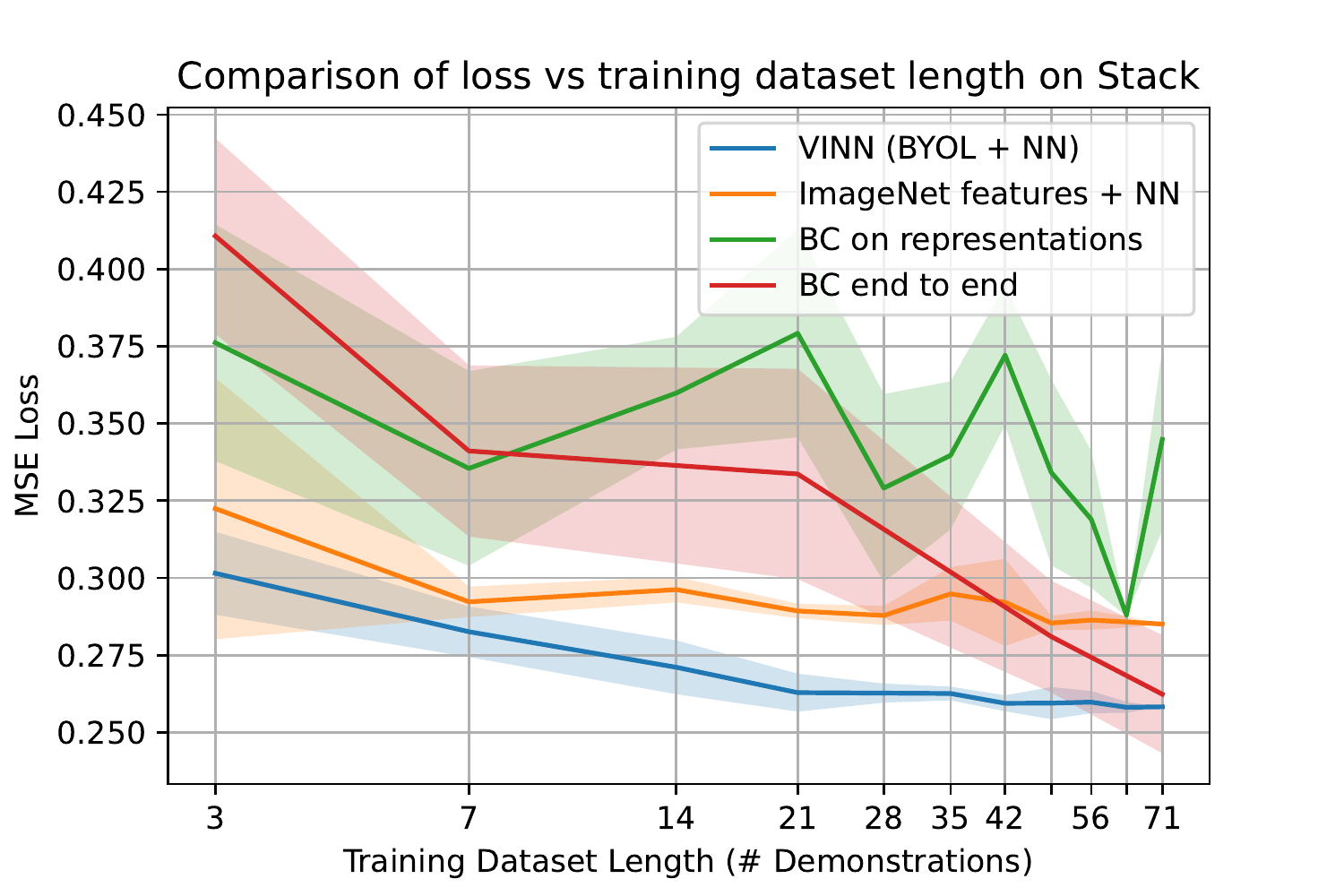}
  %\caption{A really Awesome Image}\label{fig:awesome_image2}
\endminipage\hfill
\minipage{0.33\textwidth}%
  \includegraphics[width=\linewidth]{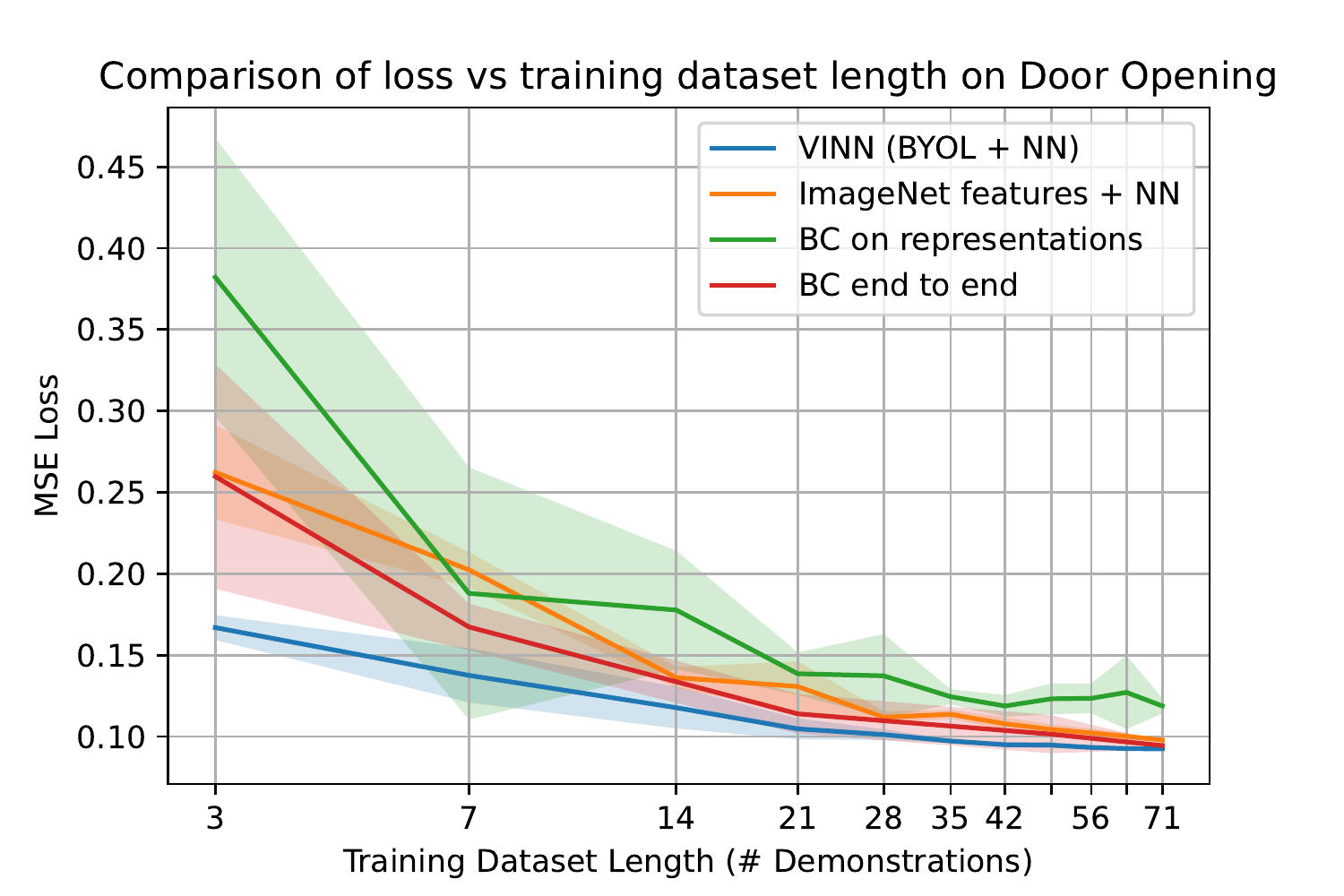}
  %\caption{A really Awesome Image}\label{fig:awesome_image3}
\endminipage
 \caption{Mean Squared Error for the Pushing, Stacking and Door Opening (left to right) datasets of different algorithms trained on subsamples of the original dataset. End-to-end behavior cloning initialized with ImageNet-trained features perform as well as VINN for larger datasets, but fixed representation based methods outperforms it largely on small datasets.}
 \label{fig:mse_loss}
\end{figure*}

% Overview of the key research questions we are looking to answer.
In the previous sections we have described our framework for visual imitation, VINN. 
In this section, we seek to answer our key question: how well does VINN imitate human demonstrations? 
To answer this question, we will evaluate both on offline datasets and in closed-loop real-robot evaluation settings.
Additionally, we will probe into the generalization with few demonstrations ability of VINN in settings where imitation algorithms usually suffer.

\subsection{Experimental Setup}\label{sec:exp_setup}
We conduct two different set of experiments: the first on the offline datasets for Pushing, Stacking and Door-Opening and the second on real-robot door opening.

\paragraph{Offline Visual Imitation Datasets} Data for Pushing 
and Stacking tasks are taken from \cite{young2020visual}. 
% We use \lpnote{Jyo} examples from the dataset provided in and use \lpnote{Jyo} of them in the training set.
The goal in the pushing task is to slide an object on a surface into a red circle.
In the stacking task, the goal is to grasp an object present in the scene and move it on top of another object also in the scene, and release. To avoid confusion, in the expert demonstrations for stacking, the closest object is always placed on top of the distant object. 
The action labels are end-effector movements, which in this case is the translation vector in between the current frame and the subsequent one. 
In each case, there are a diverse set of backgrounds and objects that make up the scene and the task, making the tasks difficult.

For Door Opening, data is collected by 3 data-collectors in their kitchens. 
This amounts to a total of 71 demonstrations for training and 21 demonstrations for testing. 
We normalize all actions from the dataset to account for scale ambiguity from SfM. 
For all three tasks, we calculate the MSE loss between the ground truth actions and the actions predicted by each of the methods. 
Note that the number of demonstrations collected for this Door Opening task is an order of magnitude smaller than the ones used for Stacking and Pushing, which contain around 750 and 930 demonstrations respectively.
To understand the performance on the various model in low data settings, we create subsampled Pushing and Stacking datasets containing 71 demonstrations on each for training and 21 for testing. This subsampling makes all three our datasets have the same size. 

\paragraph{Closed-loop control} We conduct our robot experiments on a loaded cabinet door opening task (see Fig.~\ref{fig:intro}), where the goal of the robot is to grab hold of the cabinet handle and pull open the cabinet door. 
We use the Hello-Robot Stretch~\cite{kemp2021design} for this experiment.  
% For our demonstrations, we followed the DemoAt \cite{young2020visual} method of attaching a GoPro to a Reacher-Grabber, and collecting demonstrations with it. 
% We extracted the ground truth actions from the GoPro metadata.
When evaluations start, the arm resets to $\approx0.15$ meters away from the cabinet door, with a random lateral translation within $0.05$ meters parallel to the cabinet to evaluate generalization to varying starting states.

\subsection{Baselines}
We run our experiments for baseline comparison using the following methods:
\begin{itemize}
    \item \textit{Random Action:} In this baseline, we sample a random action from the action space.
    \item \textit{Open Loop:} We find the maximum-likelihood open loop policy given all our demonstration, which is the average action $\overline a(t)$ over all actions $a_i(t)$ seen in the dataset at timestep $t$. In a Bayesian sense, if standard behavioral cloning is trying to approximate $p(a \mid s)$, this model is trying to approximate $p(a \mid t)$.
    \item \textit{Behavioral Cloning (BC) end to end:} We train a ResNet-50 model with augmentated demonstration frames similar to \cite{torabi2018behavioral,young2020visual}. We initialize the model with  weights derived from ImageNet pretraining.
    \item \textit{BC on Representations (BC-rep):} We use a self-supervised BYOL model to extract the encoding of each of our demonstration frames, and perform behavioral cloning on top of the representations.
    This baseline is similar to \cite{young2021playful} and performs better than end-to-end BC on the real robot (Table~\ref{table:real_robot}).
    \item \textit{Implicit Behavioral Cloning:} We train Implicit BC \cite{florence2021implicit} models on the tasks, modifying the official code. 
    \item \textit{ImageNet features + NN:} Instead of self-supervision, here we use the image representation generated by a pretrained ImageNet encoder akin to \cite{chen2020robust}. The difference between this baseline and our method is simply forgoing the finetuning step on our dataset.
    This baseline highlights the importance of self-supervised pre-training on the domain related dataset.
    
    \item \textit{Self-supervised learning method + NN:} This is our method; we compare three different ways of learning self-supervised representations features from our dataset -- BYOL~\cite{byol}, SimCLR~\cite{simclr}, and VICReg~\cite{bardes2021vicreg}, starting from an ImageNet pretrained ResNet-50, and then we use locally weighted regression to find the action.
    % However, the official implementation requires the action space to be products of real intervals, while in our tasks are normalized actions on the unit 2-sphere $S^2$, which is why IBC suffers in these tasks.
    % \item \textit{ImageNet Features + NN:} We use a similar approach as BC-rep, but instead of using representations trained with BYOL, we use representations generated by a ResNet50 pretrained on ImageNet only. 
    % \item \textit{Random-NN:} Similar to ImageNet-NN, we consider a randomly initialized ResNet-50 model for this baseline, get encodings from it, and get an action by performing a locally weighted $k$-NN averaging like VINN. This baseline, once again, highlights the importance of the pre-training.
    % \item \textit{BC-Random:} We use a similar approach as BC-rep, but instead of using representations from BYOL, we use representations generated by a ResNet50 pretrained on ImageNet only.
\end{itemize}

\subsection{Training Details}
Each encoder network used in this paper follows the ResNet-50 architecture~\cite{he2015deep} with the final linear layer removed. 
Unless specified otherwise, we always initialize the weights of the ResNet-50 encoder with a pretrained model on ImageNet dataset.
For VINN, we train our self-supervised encodings with the BYOL~\cite{byol} loss.
For standard end-to-end BC, we replace the last linear layer with a three-layer MLP and train it with the MSE loss.
For BC-rep, we freeze the encoding network to the weights trained by BYOL on our dataset, and train just the final layers with the MSE loss.
Additionally, for all visual learning, we use random crop, random color jitter, random grayscale augmentations and random blurring. 
We trained the self-supervised finetuning methods for 100 epochs on all three datasets. 
% What type of networks used? information to reproduce this learning portion of this paper.

% \subsection{Can VINN Compete with Standard Behavior Cloning?}
% % Experimental results on Push MSE, Stack MSE and Door MSE.
% \begin{figure}
%     \centering
%     \includegraphics[width=\linewidth]{figures/comparison_push.pdf}
%       \vspace{-0.2in}
%     % \caption{Caption}
%     %   \vspace{-0.2in}
%     \includegraphics[width=\linewidth]{figures/comparison_stack.pdf}
%       \vspace{-0.2in}
%     \includegraphics[width=\linewidth]{figures/comparison_handle.pdf}
%       \vspace{-0.1in}
%     \caption{Comparison of the performance of three candidate algorithms (BC on representations, ImageNet pretraining features + NN, and VINN (which is BYOL-fine-tuned features + NN) on the stack, push and handle datasets as the size of the dataset changes.}
%       \vspace{-0.2in}
%     \label{fig:scaling_data}
% \end{figure}
% \begin{figure}
%     \centering

% \end{figure}

\subsection{How does VINN Perform on Offline Datasets?}
For our first evaluation, we compare our method against the baselines on their Mean-Squared Error loss for the Pushing, Stacking, and Door-Opening tasks in Fig.~\ref{fig:mse_loss}. To understand the impact of the training dataset size on the algorithms, we train models on multiple subsamples of different sizes from each dataset. We see that while end-to-end Behavioral Cloning starting from pretrained ImageNet representations can be better with a large amounts of training demonstrations, Nearest Neighbor methods are either competitive or better performing in low data settings.

On the Stacking and Door-Opening tasks, VINN is significantly better when the number of training demonstrations are small ($<20$). While on the Pushing task, we notice that the task might be too difficult to solve with small number of demonstrations. One reason for this is that BYOL might not be able to extract the most relevant representations for this task. Further experiments in Table~\ref{tab:mse-table} show that using other forms of self-supervision such as VICReg can significantly improve performance on this task. Overall, these experiments supports our hypothesis that provided with good representations, nearest-neighbor techniques can provide a competitive alternative to end-to-end behavior cloning.

% it remains on par with the baselines. 
% As we can see, VINN achieves comparable MSE loss in the test set in Pushing, Stacking, and Door Opening. Thus, in simple offline datasets, the predicted action from VINN is similarly close to the ground truth as BC and BC variants. \lpnote{this statement may not be true.}

\subsection{How does VINN Perform on Robotic Evaluation?}

\begin{figure*}[t]
  \begin{center}
    \includegraphics[width = \textwidth]{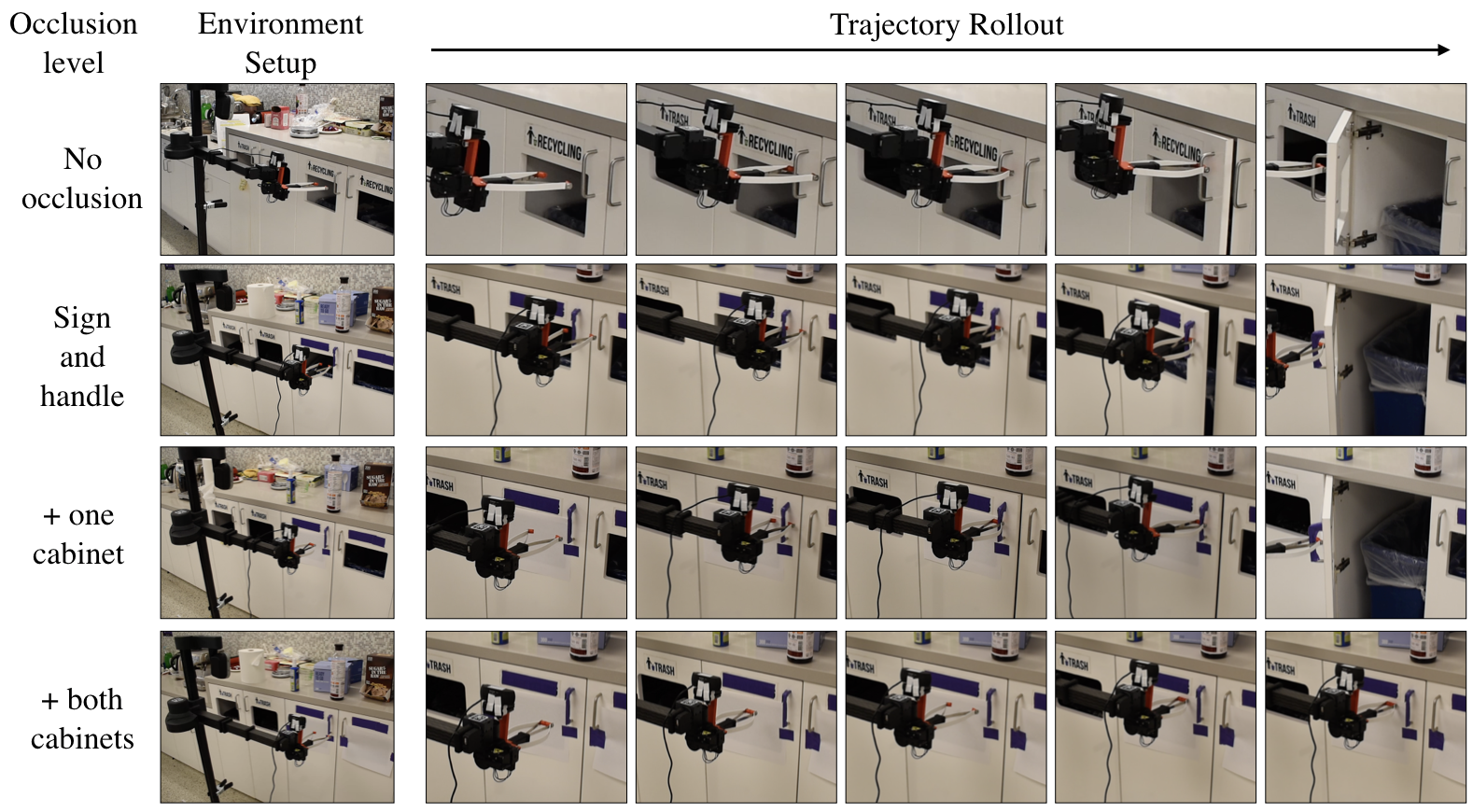}
  \end{center}
  \caption{Sample frames from the rollouts from our model on the real robot experiments, with artificial occlusions added to the cabinet to test generalization. Under the maximum occlusion, our model fails to ever open the cabinet door, while in all other cases, the robot is able to succeed (Table~\ref{table:generalization}.)}
\label{fig:rollouts}
\end{figure*}

% Experimental results on door opening without variations.
Next, we run VINN and the baselines on our real robot environment. 
In this setting, our test environment comprises of the same three cabinets where training demonstrations were collected presented without any visual modifications.
For each of our models, we run 30 rollouts with the robot in the real world with three different cabinets.
On each rollout, the starting position of the robot is randomized as detailed in (Sec.~\ref{sec:exp_setup}).
In Table~\ref{table:real_robot}, we show the percentage of success from the 30 rollouts of each model, where we record both the number of time the robot successfully grasped the handle, as well as the number of time it fully opened the door.

\begin{table}[!ht]
\centering
\caption{Success rate over 30 trials (10 trials on three cabinets each) on the robotic door opening task.}
\label{table:real_robot}
\begin{tabular}{@{}ccc@{}}
\toprule
Method                  & Handle grasped & Door opened   \\ \midrule
BC (end to end)        & 0\%            & 0\%           \\
BC on representations  & 56.7\%         & 53.3\%        \\
Imagenet features + NN & 20\%           & 0\%           \\
VINN (BYOL + NN)       & \textbf{80}\%  & \textbf{80}\% \\ \bottomrule
\end{tabular}
\end{table}

As we see from Table~\ref{table:real_robot}, VINN does better than all BC variants in successfully opening the cabinet door when there is minimal difference between the test and the train environments. 
Noticeably, it shows that depending on self-supervised features on augmented data make the models much more robust.
% We believe this success is due such models being more robust to the out of distribution queries, since the door opening task is multi-step and long horizon. 
BC, as an end-to-end parameteric model, does not have a strong prior on the actions if the robot makes a wrong move causing the visual observations to quickly goes out-of-distribution~\cite{dagger2010}. 
% However, pretrained representation based models do not overfit much to the combination of visual feature and the task objective.
% This lack of co-optimization between visual features and policy goals enforces a bottleneck that results in better visual representations in runtime.
On the other hand, VINN can recover up to certain degree of deviation using the nearest neighbor prior, since the translation actions typically tend to re-center the robot instead of pushing it further out of distribution. 
% \lpnote{Explanations in this section are quite hand-wavy. Mahi, can you edit?}

\subsection{To What Extent does VINN Generalize to Novel Scenes?}\label{sec:generalize}
% Experiments on generalization. Where does it work and where does it not.

To test generalization of our robot algorithms to novel scenes in the real world, we modified one of our test cabinets with various levels of occlusion. We show frames from a sample rollouts in each environment in Fig.~\ref{fig:rollouts}, which also shows the cabinet modifications.

\begin{table}[ht!]
\centering
\caption{Success rate over 10 trials on robotic door opening with visual modifications on one cabinet door.}
\label{table:generalization}
\begin{tabular}{@{}ccc@{}}
\toprule
Modification                               & BC-rep        & VINN (ours)    \\ \midrule
Baseline (no modifications)          & \textbf{90}\% & 80\%           \\
Covered signs and handle             & 10\%          & \textbf{70\% } \\
Covered signs, handle, and one bin   & 0\%           & \textbf{50\%}  \\
Covered signs, handle, and both bins & 0\%           & 0\%            \\ \bottomrule
\end{tabular}
\end{table}

In Table~\ref{table:generalization}, we see that VINN only completely fails when all the visual landscape on the cabinet is occluded. This failure is expected, because without coherent visual markers, the encoder fails to convey information, and thus the k-NN part also fails.
Even then, we see that VINN succeeds at a higher rate even with significant modifications to the cabinet while BC-rep fails completely. 

Over all the real robot experiments, we find the following phenomenon: while a good MSE loss is not sufficient for a good performance in the real world, the two are still correlated, and a low MSE loss seems to be necessary for good real world performance.
This observation let us test hypotheses offline before deploying and testing them in a real robot, which can be time-consuming and expensive.
We hypothesize that this gap between performance on the MSE metric (Table~\ref{tab:mse-table}) and real world performance (Table~\ref{table:real_robot}, ~\ref{table:generalization}) comes from variability in different models' ability to perform well in situations off the training manifold, where they may need to correct previous errors.
% Interestingly, although the MSE metric does not directly correspond to online performance, our experiments in Table~\ref{fig:mse_loss} suggest strong correlation. Specifically, methods that have high MSE on online tasks, mainly those that have no pretraining for representations, yield poor online performance on the Door Opening task. 

\begin{table*}[!bhtp]
\caption{\label{table:mse_loss} Test MSE $(\times 10^{-1})$ on predicted actions for a set of baseline methods and ablations. Standard deviations, when reported, are over three randomly initialized runs.}
\label{tab:mse-table}
% \resizebox{\textwidth}{!}{%
\centering
\begin{tabular}{@{}cccccccccc@{}}
\toprule
             &        &                                                     & \multicolumn{2}{c}{No Pretraining}                & \multicolumn{5}{c}{With ImageNet Pretraining}                                                                                                                                                                                                      \\ \cmidrule(l){4-5}\cmidrule(l){6-10} 
Tasks        & Random & \begin{tabular}[c]{@{}c@{}}Open\\ Loop\end{tabular} & \begin{tabular}[c]{@{}c@{}}Implicit\\ BC\end{tabular}   & \begin{tabular}[c]{@{}c@{}}BYOL\\ + NN\end{tabular} & BC-Rep          & \begin{tabular}[c]{@{}c@{}}VINN\\(BYOL + NN)\end{tabular} & \begin{tabular}[c]{@{}c@{}}VICREG\\ + NN\end{tabular} & \begin{tabular}[c]{@{}c@{}}SimCLR\\ + NN\end{tabular} & \begin{tabular}[c]{@{}c@{}}ImageNet\\ + NN\end{tabular} \\ \midrule
Door Opening & $6.34$ & $2.27$                                              & $1.8$ & $1.52$                                              & $1.19 \pm 0.05$ & $0.92$                                              & $1.05$                                                & $0.95$                                                & $0.98$                                                  \\
Stacking     & $6.13$ & $2.83$                                              & $7.1$ & $2.82$                                              & $3.45 \pm 0.29$ & $2.58$                                              & $2.74$                                                & $2.63$                                                & $2.85$                                                  \\
Pushing      & $6.15$ & $2.12$                                              & $5.6$ & $2.43$                                              & $2.20 \pm 0.20$ & $2.43$                                              & $1.50$                                                & $2.21$                                                & $2.35$                                                  \\ \bottomrule
\end{tabular}
% }
\end{table*}

\subsection{How Important are the Design Choices Made in VINN for Success?}
% In this section, we examine some of the design choices made for VINN and empirically show their impact compared to possible alter-tives. 

% \lpnote{This section needs better organization. Currently it reads like a block of text.} 
VINN comprises of two primary components, the visual encoder and the nearest-neighbor based action modules.
In this section, we consider some major design choices that we made for each of them. 

\paragraph{Choosing the Right Self-supervision} While we use a BYOL-based self-supervised encoding in our algorithm, there are multiple other self-supervised methods such as SimCLR and VICReg ~\cite{simclr,bardes2021vicreg}. On a small set of experiments we noticed similar MSE losses compared to SimCLR~\cite{simclr} and VICReg~\cite{bardes2021vicreg}. From Table~\ref{table:mse_loss}, we see that BYOL does the best in Door-Opening and Stacking, while VICReg does better in Pushing. However, we choose BYOL for our robot experiments since it requires less tuning overall. 
% \lpnote{On Pushing VICReg does better. I think it is good to highlight that k-NN depends on the quality of representations, and if it is good, it will do well.}

\paragraph{Ablating Pretraining and Fine-tuning} Another large gain in our algorithm is achieved by initializing our visual encoders with a network trained on ImageNet. In Table~\ref{table:mse_loss}, we also show MSE losses from models that resulted from ablating this components of VINN. Removing this component achieves the column BYOL + NN (No Pretraining), which performs much worse than VINN. 
Similarly, the success of VINN depends on the self-supervised fine-tuning on our dataset, ablating which results in the model shown in ImageNet + NN column of Table~\ref{table:mse_loss}. 
This model performs only slightly worse than VINN on the MSE metric.
However, in Table~\ref{table:real_robot}, we see that this model performs poorly on the real world.
These ablations show that the performance of our locally weighted regression based policy depends on the quality of the representation, where a good representation leads to better nearest neighbors, which in turn lead to a better policy both offline and online.

\paragraph{Performing Implicit instead of Explicit Imitation}
Moving away from the explicit forms of imitation where the models try to predict the actions directly, we run baselines with Implicit Behavioral Cloning (IBC)~\cite{florence2021implicit}. 
As we see on Table~\ref{table:mse_loss}, this baseline fails to learn behaviors significantly better than the random or open loop baselines. We believe this is caused by two reasons. First, the implicit models have to model the energy for the full space (action space $\times$ observation space), which requires more data than the few demonstrations that we have in our datasets. Second, the official implementation of IBC supports $[-1, 1]^3$ as the action space instead of its much smaller subspace of normalized 3d vectors $S^2$. This much larger action space, over which IBC tried to model the action, might have resulted in worse performance for IBC. While VINN makes the implicit assumption that the locally-weighted average of valid actions also yield a valid action, it can be freely projected to any relevant space without further processing, which makes it more flexible.

\paragraph{Learning a Parametric Policy on Representations}
Our Behavioral Cloning on representations (BC-Rep) baseline in all our experiments (Sec.~\ref{sec:experiments}) show the performance of a baseline where we use learned representations to learn a parametric behavioral policy. In the MSE losses (Table~\ref{tab:mse-table}) and real world experiments (Table~\ref{table:real_robot},~\ref{table:generalization}.) This is the baseline that achieves the closest performance to VINN. However, the difference between BC-rep and VINN becomes more pronounced as the gap between training and test domain or the policy horizon grows. These experimental results indicate that using a non-parametric policy may be enabling us to be robust to out-of-distribution samples.

\begin{figure}[ht]
  \begin{center}
    \includegraphics[width = \linewidth]{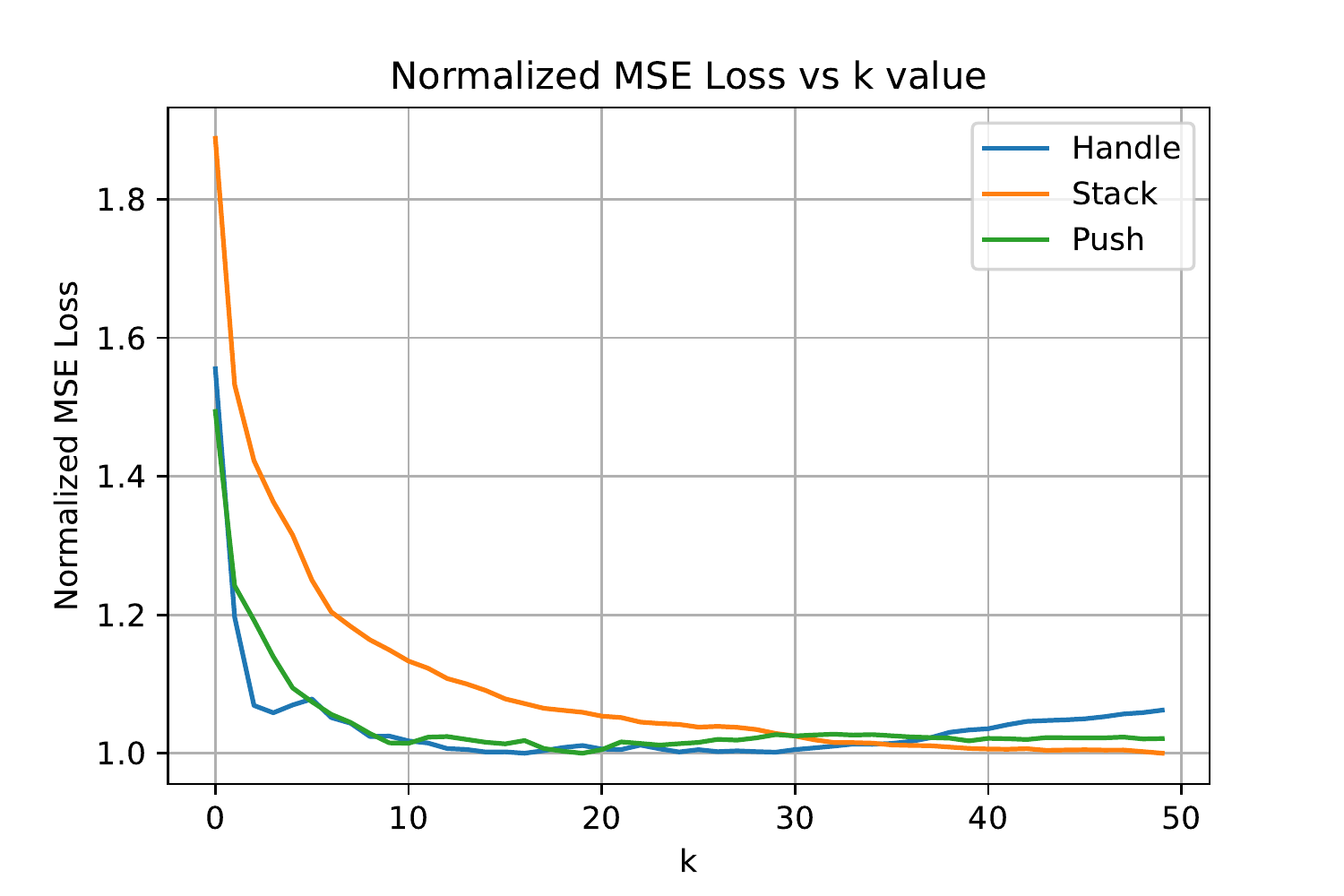}
  \end{center}
%   \vspace{-0.2in}
  \caption{Value of $k$ in the $k$-nearest neighbor weighted regression in VINN vs normalized MSE loss achieved by the model.}
%   \vspace{-0.275in}
\label{fig:knn_ablation}
\end{figure}

\paragraph{Choosing the Right $k$ for $k$-Nearest Neighbors} Finally, in VINN, we study the effect of different values of $k$ for the $k$-NN based locally weighted controller. This parameter is important because with too small of a $k$, the predicted action may stop being smooth. On the other hand, with too large of a $k$, unrelated examples may start influencing the predicted action. By plotting our model's normalized MSE loss in the validation set against the value of $k$ in Fig.~\ref{fig:knn_ablation}, we find that around $10$, $k$ seems ideal for achieving low validation loss while averaging over only a few actions. 
Beyond $k=20$, we didn't notice any significant improvement to our model from increasing $k$.

\subsection{Computational Considerations}
While the datasets we used for our experiments were not large, we recognize that our current nearest neighbor implementation is a $O(n)$ algorithm dependant linearly on the size of the training dataset with a naive algorithm. 
However, we believe VINN to be practical, since firstly, it was designed mostly for the small demonstration dataset regime where $O(n)$ is quite small, 
and secondly, this search can be sped up with a compiled index beyond the naive method using open-source libraries such as FAISS~\cite{faiss} which were optimized to run nearest neighbor search on the order of billion examples~\cite{matsui2018survey}. 
Currently, our algorithm takes $\approx 0.074$ seconds to encode an image, and $\approx 0.038$ seconds to perform nearest neighbors regression, which is only a small speed penalty for the robotic tasks we consider.
% \lpnote{Can we say that more efficient methods do exist for faster retrieval? and cite them as well.}

%%%%%%%%%%%%%%%%%%%%%%%%%%%%%%%%%%%%%%%%%%%%%%%%%%%%%%%%%%%%%%%%%%%%%%%%%%%%%%%%
\section{Limitations and Future Work}
% \lpnote{First line should be summary of what this paper has presented. Maybe not call it clear drawbacks?}
In this work we proposed VINN, a new visual imitation framework that decouples visual representation learning from behavior learning.
Although this decoupling improves over standard visual imitation methods, there are several avenues for future work.
First, there is still some remaining hurdles to generalizing to a new scene, as seen in Sec.~\ref{sec:generalize}, where our model fails when all large, recognizable markers are removed from the scene.
While our NN-based action estimation lets us add new demonstrations easily, we cannot easily adapt our representation to such drastic changes in scene.
An incremental representation learning algorithm has great potential to improve upon that. 
Second, our self-supervised learning is currently done on task related data, while ideally, if the dataset is expansive enough, task agnostic pre-training should also give us good performance ~\cite{young2021playful}. 
Finally, although our framework focuses on a single-task setting, we believe that learning a joint representation for multiple tasks could reduce the overall training overhead while being just as accurate. 

%%%%%%%%%%%%%%%%%%%%%%%%%%%%%%%%%%%%%%%%%%%%%%%%%%%%%%%%%%%%%%%%%%%%%%%%%%%%%%%%
\section*{Acknowledgements}
We thank Rohith Mukku for his help with writing code and running ablation experiments. We thank Dhiraj Gandhi, Pete Florence, and Soumith Chintala for providing feedback on an early version of this paper. This work was supported by grants from Honda, Amazon, and ONR award numbers N00014-21-1-2404 and N00014-21-1-2758.

%%%%%%%%%%%%%%%%%%%%%%%%%%%%%%%%%%%%%%%%%%%%%%%%%%%%%%%%%%%%%%%%%%%%%%%%%%%%%%%%
% \section*{ACKNOWLEDGEMENTS}

%% Use plainnat to work nicely with natbib. 

\bibliographystyle{plainnat}
\bibliography{references}

\clearpage

\appendix
\subsection{VINN Pytorch Pseudocode}
% \textbf{VINN Pytorch Pseudocode}
\begin{lstlisting}
def dist_metric(x,y):
    return(torch.norm(x-y).item())

def calculate_action(dist_list,k):
    action = torch.tensor([0.0,0.0,0.0])
    top_k_weights = torch.zeros((k,))
    for i in range(k):
        top_k_weights[i] = dist_list[i][0]
    top_k_weights = softmax(-1*top_k_weights)
    for i in range(k):
        action = torch.add(top_k_weights[i] 
            * dist_list[i][1], action)
    return(action)

def calculate_nearest_neighbors(query_img, dataset, k):
        query_embedding = encoder(query_img)
        for dataset_index in range(len(dataset)):
            dataset_embedding, dataset_translation = dataset[dataset_index]
            distance = dist_metric(query_embedding, dataset_embedding)
            dist_list.append((distance, dataset_translation, dataset_path))
       dist_list = sorted(dist_list, key = lambda tup: tup[0])
pred_action = calculate_action(dist_list, k)
return pred_action

\end{lstlisting}

\subsection{Network Architectures and Training Details}
In this section, we will go over our implementation, network architectures, and training details for our various baselines.

\paragraph{Random Action} We sampled a 3-d vector from $[-1, 1]^3$, normalized it, and used it as our action for this baseline.

\paragraph{Open Loop} We computed the average action at frame $t$ over all demonstrations from our dataset for this baseline for each frame number $t$.

\paragraph{Behavioral Cloning (end to end or from representations)}
For our parameterized model experiments, our encoding network is always a ResNet50, and our translation neural network is a three-layer MLP whose layer dimensions are $2048,1024,3$. 
Our gripper model is a linear layer that predicts four gripper states from the encoder network output.
We train both models for 8000 epochs with a learning rate of 0.001 using the Adam optimizer. 
On an RTX8000, given the learned representations, it takes 12 minutes on the Door Opening dataset to train both MLPs for BC from representations.
For training the BC end to end model until convergence, it takes us three hours in total.

\paragraph{Implicit Behavioral Cloning (IBC)} We used the official Github repo for Implicit Behavioral Cloning~\cite{florence2021implicit} offline experiments. We modified their Push from Pixels task to fit 3-d vectors bounded within $[-1, 1]^3$. Unfortunately, we could not use the space of normal vectors since the current published version of the IBC code does not support constrained action spaces.

We trained the standard Dense ResNet model provided with the IBC repo for encoders, and IBC-with-DFO framework for sampling actions. It took us about 6 hours to train the models end-to-end on our datasets on an RTX 8000 GPU for 10,000 steps. For every hyperparameter, we use the defaults for the learning to push from pixels task that is included in the IBC repository.

We computed the MSE loss from this model by first sampling 256 actions with DFO optimization, as it's done in IBC, and choosing the action with the highest assigned value out of it.

\paragraph{VINN}
For our BYOL-trained encoding network, we use a ResNet50 architecture, with the final linear for ImageNet classification replaced with an identity layer. We use the representation vector of size 2048. We fine-tune this network using BYOL for a 100 epochs on our demonstration datasets with the ADAM optimizer and a $3\times 10^{-4}$ learning rate. To train this BYOL on the Door Opening dataset for 100 epochs it took approximately 3.5 hours on a workstation with one Nvidia RTX8000. 

\subsection{Robot details}
\begin{figure}[!ht]
    \centering
    \includegraphics[width=0.75\linewidth]{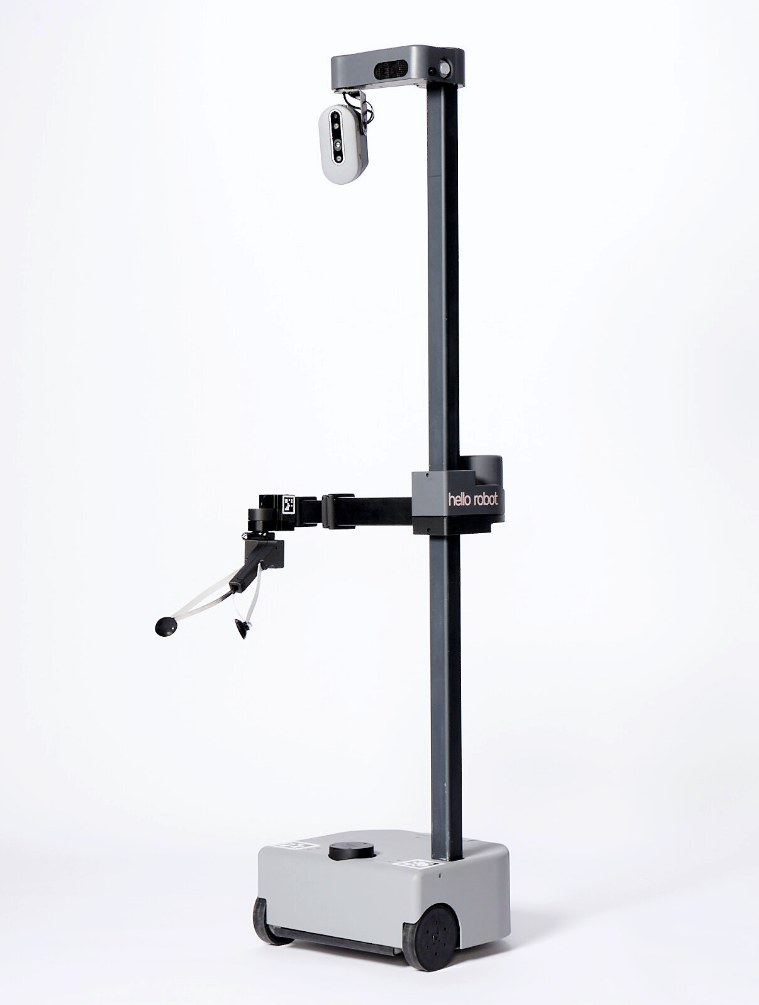}
    \caption{Hello Robot's Stretch~\cite{kemp2021design}, the robot model used in our experiments}
    \label{fig:reacher}
\end{figure}

We run all of our robots in the Hello Robot's Stretch~\cite{kemp2021design}. This robot has a dexterous wrist with 3 DoF, a telescopic 1 DoF arm on which it is mounted, and an 1-DoF lift on which the arm is mounted. The base of the robot is also capable of rotation and lateral motion, which gives the robot's end-effector a full 6-DoF capability. 

On each step, the translation model predicts $\Delta(x,y,z)$ for the gripper, which is converted to the movement in the robot's joints with an inverse kinematics model. This model takes into account simpler objectives like avoiding self-collisions, but does not model avoiding issues like environment collisions.

For the robot observations, we use a standard webcam mounted on top of the robot wrist using a custom 3-d printed mount. The image captured by the robot is streamed over the network to a machine running the VINN algorithm, which responds with the predicted robot action.

\subsection{Demonstration Collection Details}
\begin{figure}[!ht]
    \centering
    \includegraphics[width=\linewidth]{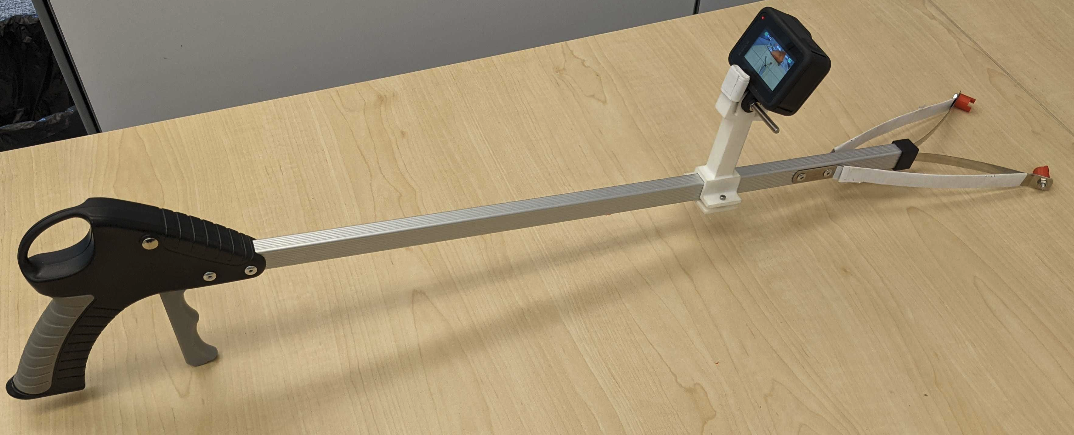}
    \caption{Reacher grabber tool used for our demonstrations.}
    \label{fig:reacher}
\end{figure}

We use the DemoAT~\cite{young2020visual} framework for collecting our demonstrations. We use a simple reacher-grabber tool availabe at hardware shops or online, fitted with a GoPro camera to do capture our observation frames. An image of this is shown in Fig.~\ref{fig:reacher},

\begin{figure}[!ht]
    \centering
    \includegraphics[width=\linewidth]{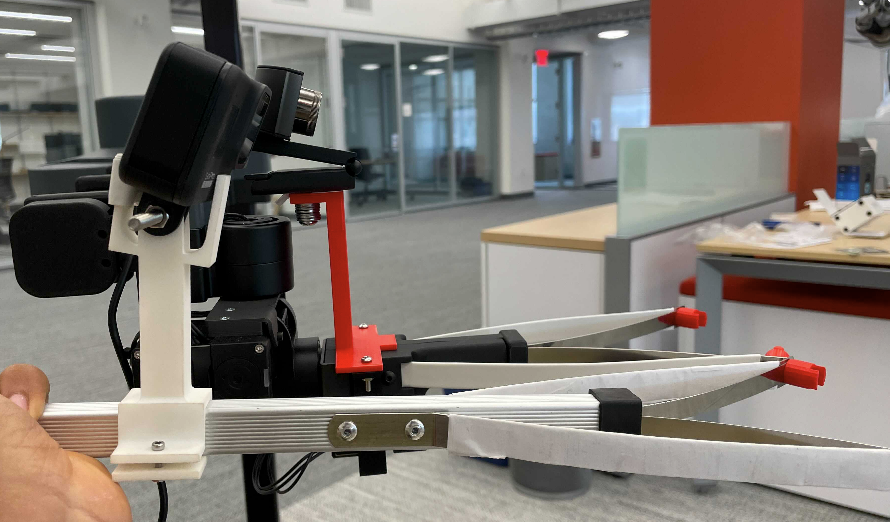}
    \caption{Modified grip on the robot and the reacher grabber.}
    \label{fig:reacher_grip}
\end{figure}

We replaced the pads at the end of the robot gripper with simple 3-d printed nubs for easy resets of the robot, and we do the same on the reacher-grabber tool, as seen in Fig.~\ref{fig:reacher_grip}.

To get visual observations, we mount a GoPro on top of the reacher grabber tool with a custom 3-d printed mount. We linearize the GoPro video in post-processing using \texttt{ffmpeg} to get rid of the wide-angle distortions, and extract the frames at one frame per second speed. Finally, using the extracted frames and the OpenSfM library, we reconstruct the 3-d movements between frames. We take the delta position changes between consecutive frames, and normalize them to get our actions.

\end{document}